\def\year{2021}\relax
\documentclass[letterpaper]{article} 
\usepackage{aaai21}  
\usepackage{times}  
\usepackage{helvet} 
\usepackage{courier}  
\usepackage[hyphens]{url}  
\usepackage{amsmath} 
\usepackage{amssymb}
\usepackage{multirow}
\usepackage{graphicx} 
\usepackage{natbib}  
\usepackage{caption} 
\title{Aggregated Multi-GANs for Controlled 3D Human Motion Prediction}
\author{
    Zhenguang Liu\textsuperscript{\rm 1},
    Kedi Lyu\textsuperscript{\rm 2}\thanks{Corresponding Authors},
    Shuang Wu\textsuperscript{\rm 3*},
    Haipeng Chen\textsuperscript{\rm 2*},
    Yanbin Hao\textsuperscript{\rm 4},
    Shouling Ji\textsuperscript{\rm 5}
    \\
}
\affiliations{
   \textsuperscript{\rm 1}Zhejiang Gongshang University\\
   \textsuperscript{\rm 2}Jilin University\\
   \textsuperscript{\rm 3}Nanyang Technological University\\ \textsuperscript{\rm 4}University of Science and Technology of China\\
    \textsuperscript{\rm 5}Zhejiang University\\
    {\textrm\small liuzhenguang2008@gmail.com},
    {\textrm\small lvkd19@mails.jlu.edu.cn},
    {\textrm\small wushuang@outlook.sg},
    {\textrm\small chenhp@jlu.edu.cn},
    {\textrm\small haoyanbin@hotmail.com},
    {\textrm\small sji@zju.edu.cn}
}

\begin{document}

\maketitle

\begin{abstract}
Human motion prediction from historical pose sequence is at the core of many applications in machine intelligence. However, in current state-of-the-art methods, the predicted future motion is confined within the same activity. One can neither generate predictions that differ from the current activity, nor manipulate the body parts to explore various future possibilities. Undoubtedly, this greatly limits the usefulness and applicability of motion prediction. In this paper, we propose a generalization of the human motion prediction task in which control parameters can be readily incorporated to adjust the forecasted motion. Our method is compelling in that it enables manipulable motion prediction across activity types and allows customization of the human movement in a variety of fine-grained ways. To this aim, a simple yet effective composite GAN structure, consisting of local GANs for different body parts and aggregated via a global GAN is presented. The local GANs game in lower dimensions, while the global GAN adjusts in high dimensional space to avoid mode collapse. Extensive experiments show that our method outperforms state-of-the-art. The codes are available at https://github.com/herolvkd/AM-GAN.
\end{abstract}

\section{Introduction}
\begin{figure}[ht]
	\centering
	\includegraphics[width=1\linewidth]{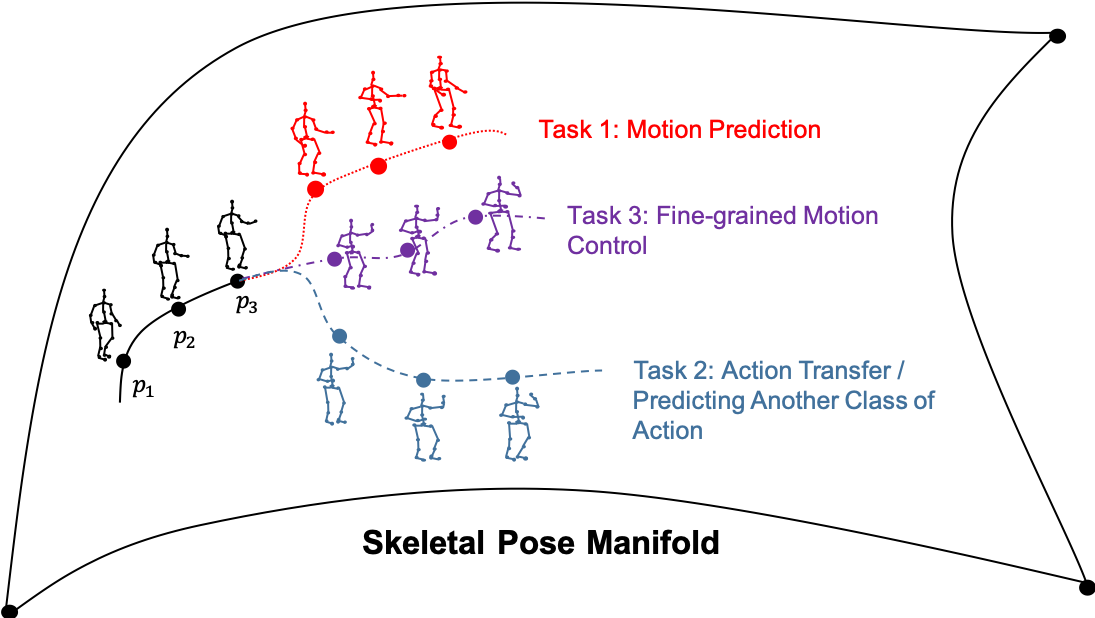}
	\caption{Problem illustration. Mathematically, an observed motion sequence traces out a trajectory on the skeletal pose manifold. The problem of human motion prediction amounts to extrapolating this trajectory in a smooth and coherent manner. We are interested in tasks: 1) predicting the most likely motion (e.g., continuation of the walking action); 2) predicting the transition to another specified action (e.g. transition from current action to eating); 3) fine-grained control over body parts, e.g., requiring the upper body performing eating action while the lower body engaging in walking.}
	\label{fig:problem}
\end{figure}

An important component of our capacity to interact with the external world resides in the ability to predict the future \cite{residualgru}, based on existing cues and past experience. Having a mental representation of how other humans are likely to act  is crucial for us to avoid dangers  and make decisions. Likewise, the ability for machines to anticipate and model human motion dynamics is very much coveted. Modeling and predicting human motion is at the heart of many important applications in the computer vision domain such as video making, human posture tracking \cite{taylor2010dynamical}, as well as the robotics domain such as regulating a robot's response and behaviour in human-machine interactions \cite{koppula2013anticipating, koppula2016anticipatory}.

Human motion prediction(HMP) is challenging since humans do not follow fixed laws of motion as is the case for inanimate objects. The inherent stochastic nature of human behaviour invokes complexity in the form of non-linearity and high dimensionality. Earlier approaches have typically utilized shallow latent-variable models such as hidden Markov models \cite{Markov} or Gaussian processes \cite{gaussian} to represent the motion dynamics with hidden states. In recent years, many deep learning approaches were proposed and have yielded improved results. 
A first class of work utilised recurrent neural networks (RNNs) such as Long Short Term Memory (LSTM) \cite{erd}, Gated Recurrent Unit (GRU) \cite{residualgru,pavllo2019modeling} or refined RNN architectures tailored for modeling motion dynamics. Another line of work built upon the success of deep generative models including Variational Autoencoders (VAEs) \cite{habibie2017recurrent, yan2018mt} and Generative Adversarial Networks (GANs) \cite{barsoum2018hp, kundu2019bihmp}.

A major shortcoming of prior works 
is that they can only predict a single output sequence for any given observed motion sequence. However, given our incomplete knowledge of the present, the future should not be represented as a single deterministic state but rather a spectrum of plausible outcomes. Considering only a single future outcome may be inadequate and lead to misinformed decision making. For example, in safety-crucial applications such as autonomous driving, unexpected maneuvers such as sudden overtaking must be accounted for in guiding the course of actions.

Another severe limitation of existing works is that they can only predict one single activity belonging to the same activity as the observed sequence, e.g. predicting `walking' motion from an observed `walking' sequence. For real applications such as video synthesis, animation, and game character movement generation, we usually want to (1) customize a future motion with fine-grained control on the body parts, e.g., requiring the whole body act in ‘walking’ activity but with the right arm act in ‘eating’, and (2) command the prediction to smoothly transfer from current activity to another activity. However, existing methods clearly fail to fulfill the above requirements. This motivates us to propose the controlled motion prediction problem.  
The ability to control and fine-tune the generated future motion would allow for a broader research space in human-machine interaction and would certainly be desirable for many applications. 

Inspired by a divide and conquer strategy, we propose a novel Aggregated Multi-GAN (AM-GAN) framework for the task of controlled human motion prediction. The generator module in our architecture allows for conditional motion generation, which may be used to realize action transfer, \emph{i.e.,} the predicted motion can be controlled to transfer to a specified action instead of continuing the original action. AM-GAN models the motion of central spine and four limbs with separate GANs and the final complete motion is aggregated over them. The aggregation process ensures that the overall dynamics and interaction between kinematic chains are balanced and attuned. Dividing the generative modeling of the entire human skeletal motion into sub kinematic chains reduces the complexity of high dimensional motion generation into a less complicated low dimensional one. Furthermore, such an approach also enables us to acquire fine-grained control over the generated motion in body chains.

To summarize, our main contributions are: 1) We propose a generalized 3D human motion prediction problem that allows for adequate and fine-grained control over the predictive spectrum. 2) We design a novel Aggregated Multi-GAN (AM-GAN) framework to tackle this problem via a divide and conquer strategy. 3) AM-GAN sets the new state-of-the-art performance for quantitative short-term prediction and also generates natural and consistent long-term prediction. More importantly, AM-GAN also allows control over the prediction such as generating  transitions to specified actions and fine-grained control of body part movements. The AM-GAN is, to our knowledge, the first work that can handle these tasks simultaneously (short-term prediction, long-term prediction, specified action transfer, and fine-grained motion prediction control).

\section{Related Work}
\textbf{RNN based motion prediction.} Deep learning has achieved great success in various fields \cite{kingma2014semi,butepage2017deep,Multiview,Leizhu,Ningxu}. RNN is a natural candidate for modeling human motion sequence due to their suitability in handling time series. Each time step of  a motion sequence is encoded as a hidden state that learns the dynamics of the motion up to the current. The final hidden state corresponding to the last observed frame is then recurrently updated to generate a future sequence. This approach has been adopted in several works \cite{erd, residualgru}. A similar work \cite{li2018convolutional} employs convolutional layers instead of RNN units. In general, the progress within this line of work seeks to better model long-term temporal dependency and overcome the issue that the current hidden state tends to be overwhelmed by the last observed frame. \cite{residualgru} proposes to incorporate a residual connection between the RNN units while \cite{liu2019towards} proposes a hierarchical RNN architecture. 
The major shortcoming of this class of works is that the output is fixed to be a single deterministic future motion sequence, and would be inadequate for obtaining a representation of the spectrum of future possibilities.\\
\hspace*{0.3cm}\textbf{GAN based human motion prediction.} GAN \cite{goodfellow2014generative} provides an alternative to the RNN based approach for HMP. In an adversarial process, a generator network outputs possible future motion sequences conditioned on an observed input sequence and a discriminator network assesses how likely these generated sequences are real or fake. In contrast to RNN-based approaches, GAN-based approaches \cite{barsoum2018hp, kundu2019bihmp} for HMP are probabilistic in that the learned generative model can output different sequences of possible future human poses from the same input sequence by sampling an additional latent noise vector. A challenging problem with GAN approaches is mode-collapsing during training, which leads to generated outputs all lying within certain modes. \cite{kundu2019bihmp} proposed a modification of the discriminator so as to incorporate both adversarial loss as well as content loss on the prediction. \\
\hspace*{0.3cm}\textbf{Controllable motion prediction.} To our knowledge, few existing works assimilate control for motion prediction tasks. \cite{holden2016deep, pavllo2019modeling} incorporate control parameters in the form of path trajectories along which a predicted walking or running action is bound to traverse. Another related work \cite{holden2017fast} adapts the technique of neural style transfer for motion generation. For example, a walking motion sequence may be manifested in a jubilant fashion or a depressed manner and these different styles can be transferred onto a walking sequence without altering the essential content. 
However, the problem of generating transitions between different class of actions has not yet been studied in the setting of HMP. Different from previous works, our approach does not focus on controlling the trajectory or expression style of a walking action. Rather, we seek to allow for predicting transitions between different actions and fine-grained control over body parts in executing a prediction.

\section{Our Approach}
\begin{figure*}[ht]
	\centering
	\includegraphics[width=0.55\textwidth]{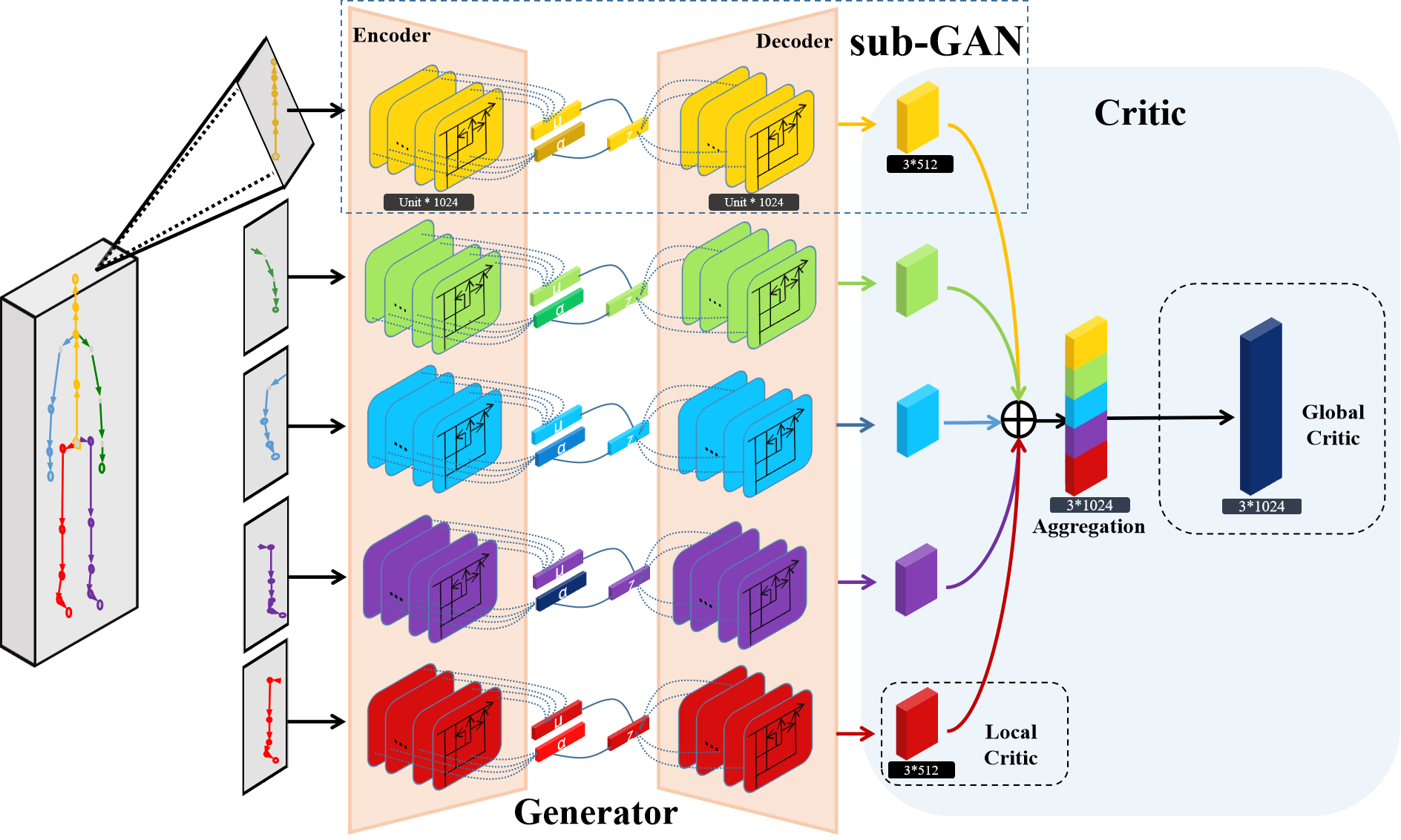}
	\caption{The proposed AM-GAN architecture. Each kinematic chain is modeled by its own sub-GAN (with matching colors). Each sub-GAN has a generator in the form of a VAE network as well as a local critic. The combined output from all sub-GANs is then aggregated and passed to a global critic. The aggregation layer and the global critic constitute the  global GAN.}
	\label{fig:architecture}
\end{figure*}
In this section, we present the details of our proposed AM-GAN. The overview of our approach is inspired by the divide and conquer strategy. We observe that by reducing the general motion generation task entailing of the entire human skeleton to its constituent skeletal kinematic chains, the issue of mode collapsing can be effectively alleviated, leading to more accurate and diverse motion sequence forecasts.

\subsection{Problem Formulation}
Presented with an observed human motion sequence parameterized as $\left(\mathbf{p}_{1},\mathbf{p}_{2},\cdots,\mathbf{p}_{t}\right)$ with optional control parameters, we are interested in generating a future pose sequence $\left(\mathbf{p}_{t+1},\mathbf{p}_{t+2},\cdots,\mathbf{p}_{t+T}\right)$.

Each human skeletal pose $\mathbf{p}_i$ is composed of distinct key joints. 
For simplicity, we adopt 3D coordinates to characterize the locations of each joint. Mathematically, the skeletal human pose lies on a manifold, and  a motion sequence traces out a trajectory on this pose manifold, which is  depicted in Fig. \ref{fig:problem}.  The problem of HMP is to find a smooth extension to the observed trajectory, while action transfer is to find a smooth transfer from current trajectory to the target one. 

\subsection{Method Overview and Rationale}
\textbf{Divide and conquer with multiple GANs and aggregation.} Existing methods attempt to model the dynamics of entire human skeletal motion in its entirety. \cite{srnn, li2018convolutional} further seek to model spatial dependencies between joints. These attempts set the problem of motion generation on an unnecessarily high dimensional manifold. Optimizing for both the adversarial and content loss on a high dimensional manifold is harder and more likely to incur issues of non-convergence, gradient diminishing and mode collapse \cite{kodali2017convergence}.

Taking into account the anatomical separability of different chains and the difficulty of motion generation in high dimensional manifolds, we propose to divide the motion generation problem into that of the constituent chains and conquer the complete problem through an aggregation process that takes into account the interaction between these chains.

Corresponding to each pose $\mathbf{p}_i$, the constituent chains will be denoted as $\mathbf{c}_i^j$ where $1\leq j \leq 5$ is the chain index (torso and four limbs). Note that we also experimented with other body divisions such as dividing each limb into two parts, but with no significant performance gain obtained.

\textbf{Controlled prediction across action types.} It is not realistic to confine human motion prediction within a single class of action, as is the case for existing methods. We propose a method that explicitly allows for predictions where one action may transit and evolve into another specified action. Therefore, instead of purely RNNs, we engage generative modeling which are better suited for this objective.

\textbf{Fine-grained control of body parts.} Previous motion prediction works are lacking in their abilities to control individual body parts in the predicted motion. We attempt to attain fine-grained control of body components and again, modeling each kinematic chain with its own specific GAN plays a major role in this. The capacity to control fine details of body parts in motion is undoubtedly a very desirable influence in fostering the versatility and usability in human motion prediction research. For example, different stylistic peculiarities or gait patterns across different individuals maybe directly integrated into the final prediction.

\begin{table*}[t]
	\centering{
	\begin{tabular}{l|l|l|l|l|l|l|l|l|l|l}
		\hline
		\multirow{2}{*}{Methods \textbackslash Time (ms)}
		& \multicolumn{5}{c|}{Walking}     & \multicolumn{5}{c}{Eating}      \\ \cline{2-11} 
		& 80 & 160 & 320 & 400 & 1000 & 80 & 160 & 320 & 400 & 1000 \\ 
		\hline
		ERD \cite{erd} & 0.93 & 1.18 & 1.59 & 1.78 & 2.24   & 1.27 & 1.45 & 1.66 & 1.80 & 2.02   \\ 
		SRNN \cite{srnn}& 0.81 & 0.94 & 1.16 & 1.30 & 1.80   & 0.97 & 1.14 & 1.35 & 1.46 & 2.11 \\
		RRNN \cite{residualgru} & 0.33 & 0.56 & 0.78 & 0.85 & 1.14 & 0.26 & 0.43 & 0.66 & 0.81 & 1.34 \\ 
		LSTM-AE \cite{tu2018spatial} & 1.00 & 1.11 & 1.39 & ----- & 1.39 & 1.31 & 1.49 & 1.86 & ----- & 2.01 \\
		CEM \cite{li2018convolutional} & 0.33 & 0.54 & 0.68 & 0.73 & 0.92 & 0.22 & 0.36 & 0.58 & 0.71 & 1.24 \\ 
		HP-GAN \cite{barsoum2018hp} & 0.95 & 1.17 & 1.69 & 1.79 & 2.47 & 1.28 & 1.47 & 1.70 & 1.82 & 2.51 \\ 
		SKelNet \cite{guo2019human} & 0.34 & 0.52 & 0.69 & 0.70 & 0.90 & 0.23 & 0.39 & 0.50 & 0.71 & 1.23 \\ 
		BiHMP-GAN \cite{kundu2019bihmp} & 0.33 & 0.52 & 0.63 & 0.67 & 0.85 & \textbf{0.20} & 0.33 & 0.54 & 0.70 & 1.20 \\
		HMR \cite{liu2019towards} & 0.34 & 0.52 & 0.67 & 0.69 & 0.90 & 0.22 & \textbf{0.31} & 0.51 & 0.69 & 1.21 \\ 
		QuaterNet \cite{pavllo2019modeling} & \textbf{0.23} & \textbf{0.37} & \textbf{0.59} & 0.66 & 0.86 & \textbf{0.20} & 0.33 & 0.54 & 0.68 & 1.17 \\ 
		AM-GAN (Ours) & \textbf{0.23} & 0.51 & 0.62 & \textbf{0.66} & \textbf{0.84} & \textbf{0.20} & \textbf{0.31} & \textbf{0.49} & \textbf{0.66} & \textbf{1.15} \\ \hline
		\multirow{2}{*}{Methods \textbackslash Time (ms)}
		& \multicolumn{5}{c|}{Smoking}     & \multicolumn{5}{c}{Discussion}  \\ \cline{2-11} 
		& 80 & 160 & 320 & 400 & 1000 & 80 & 160 & 320 & 400 & 1000 \\ \hline
		ERD \cite{erd} & 1.66 & 1.95 & 2.35 & 2.42 & 3.14 & 2.27 & 2.47 & 2.68 & 2.76 & 3.11   \\ 
		SRNN \cite{srnn} & 1.45 & 1.68 & 1.94 & 2.08 & 2.57 & 1.22 & 1.49 & 1.83 & 1.93 & 2.19   \\ 
		RRNN \cite{residualgru} & 0.35 & 0.64 & 10.3 & 1.15 & 1.83 & 0.37 & 0.77 & 1.06 & 1.10 & 1.79 \\ 
		LSTM-AE \cite{tu2018spatial} & 0.92 & 1.03 & 1.15 & ----- & 1.77 & 1.11 & 1.20 & 1.38 &-----& 1.73 \\ 
		CEM \cite{li2018convolutional} & 0.26 & 0.49 & 0.96 & 0.92 & 1.62 & 0.32 & 0.67 & 0.94 & 1.01 & 1.86 \\ 
		HP-GAN \cite{barsoum2018hp} & 1.71 & 1.89 & 2.33 & 2.42 & 3.20 & 2.29 & 2.61 & 2.79 & 2.88 & 3.67 \\ 
		SKelNet \cite{guo2019human} & 0.26 & 0.47 & 0.91 & 0.89 & 1.56 & 0.29 & 0.62 & 0.88 & 1.00 & 1.68 \\ 
		BiHMP-GAN \cite{kundu2019bihmp} & 0.26 & 0.50 & 0.91 & \textbf{0.86} & 1.11 & 0.33 & 0.65 & 0.91 & 0.95 & 1.77 \\ 
		HMR \cite{liu2019towards}  & 0.26 & 0.47 & 0.90 & 0.91 & 1.49 & 0.29 & \textbf{0.55} & 0.83 & 0.94 & 1.61 \\ 
		QuaterNet \cite{pavllo2019modeling} & 0.26 & 0.49 & 0.92 & 0.89 & 1.67 & \textbf{0.28} & 0.62 & 0.87 & 0.95 & 1.89 \\ 
		AM-GAN (Ours) & \textbf{0.25} & \textbf{0.46} & \textbf{0.88} & 0.88 & \textbf{1.10} & \textbf{0.28} & \textbf{0.55} & \textbf{0.81} & \textbf{0.92} & \textbf{1.58} \\ \hline
	\end{tabular}
	}
	\caption{Performance evaluation (in MAE) of comparison methods over 4 different action types on the H3.6m dataset}
	\label{tab:allmethods}
\end{table*}

\subsection{Architecture Overview}
The overall framework of our method is illustrated in Fig. \ref{fig:architecture}. There are five sub-GANs (local GANs), each taking care of a kinematic chain. The sub-GANs will now model motion generation on a lower dimensional manifold which is advantageous from a computational perspective and may be favorable in avoiding mode collapse. After a local discrimination process, the outputs from all sub-GANs are then combined via an aggregation   layer. The output of the aggregation layer is then passed to the global critic for assessing the quality of the generated motion sequence in its entirety. 

The conditional probability distributions learnt through the generative modeling process may then be invoked to realize action transfer in motion prediction. The sub-GANs in the AM-GAN network can further be mobilized to control the skeletal pose at a fine-grained scale. 


{\bf Generator.} In order to effectively interpolate between latent representations for generating transitions across different action types, we have adopted a Variational Auto-encoder (VAE) architecture \cite{kingma2014semi}
as the generator module for each of our sub-GAN. Since the task entails sequential timeseries data, the encoder and decoder engage GRU for modeling the motion sequences. 

Formally, the encoder (in sub-GAN $j$) maps the observed sequence $\mathbf{C}^j=\left(\mathbf{c}_1^j,\mathbf{c}_2^j,\cdots,\mathbf{c}_t^j\right)$ to a latent variable $\mathbf{z}^j$.
\begin{equation} \label{eqn:encoder}
\begin{aligned}
\mathbf{z}^j&=& \text{EN} (\mathbf{C}^j)
\end{aligned}
\end{equation}
The decoder then recursively generates a future prediction sequence based on this latent vector $\mathbf{z}^j$.
\begin{equation} \label{eqn:decoder}
\begin{aligned}
\mathbf{C}'^j&=& \text{DE} (\mathbf{z}^j)
\end{aligned}
\end{equation}

\begin{figure*}[t]
	\centering
	\includegraphics[width=0.8\linewidth]{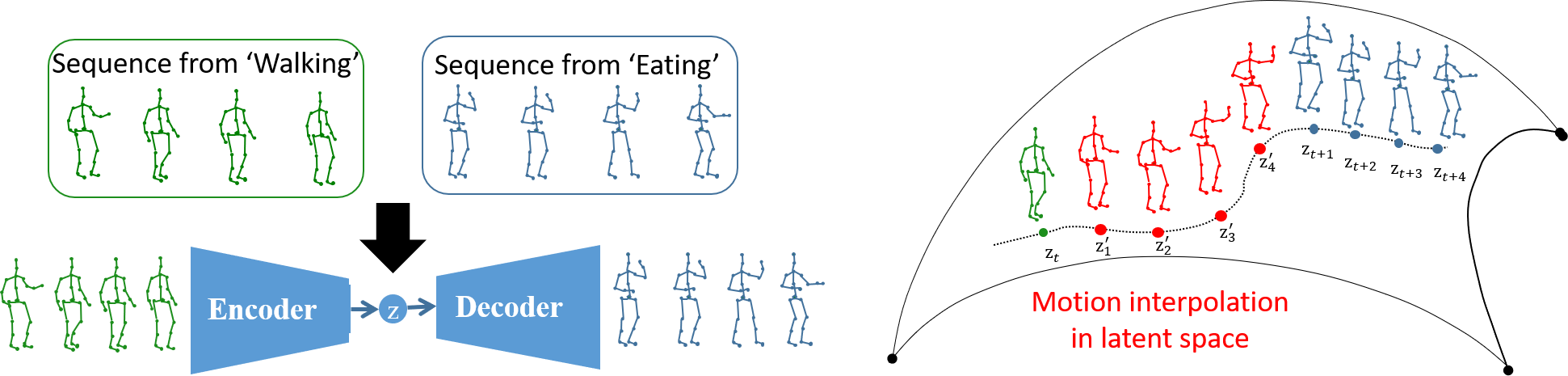}
	\caption{We perform motion interpolation in the latent space (i.e., $z$ in VAE) and insert these motion frames (red) to fill up the discontinuity between two action sequences. Different from other two tasks, here the AM-GAN network is trained to predict an immediate future pose rather than a sequence.}
	\label{fig:mergeaction}
\end{figure*}

{\bf Aggregation.} Dynamically, there is interconnection and interdependence between different kinematic chains, as typified in the synchronized movements of the right arm and left leg in walking. Therefore, the individual latent motion contexts from each kinematic chain is mapped through a fully connected aggregation layer, to the final predicted pose sequence output as follows:
\begin{equation} \label{eqn:aggregation}
\begin{aligned}
\mathbf{P}'&=&\text{Agg}\left(\sum_{j=1}^5 \mathbf{C}'^j\right)
\end{aligned}
\end{equation}
where $\text{Agg}$ denotes the mapping function of the aggregation layer. $\mathbf{P}'$ is shorthand for the future pose sequence. 

{\bf Critic.} The critic module consists of Local Critics for each sub-GAN $\text{D}^j$ and a Global Critic $\text{D}^G$ after the aggregation layer. A three-layers fully connected feedforward network is used in both types of critics. The local critics judge the generation of the five branches of the human body to ensure the accuracy of local prediction. The rationale is to first focus on the local context and reduce interference between local components or body parts at the base level. However, we cannot simply ignore the correlation and synchronization between different body parts in executing a motion. We also include a global critic to judge the feasibility of the motion and to ensure the naturalness and accuracy of the entire pose.

\subsection{Loss Function} \label{sec_lossfunction}
Now, we zoom in on the loss function for training AM-GAN. To account for chain-level and pose-level accuracy, we consider both local loss and global loss. We further engage a stability loss at the chain-level and a consistency loss at the pose-level to ensure the authenticity and temporal smoothness of the motion.

{\bf Chain-level loss.} Within each sub-GAN, a generator outputs the motion prediction of a kinematic chain $j$. A critic network $\mathbf{D}^j$ learns to assign higher scores to real future sequences and lower scores to the generated ones. This gives the Wasserstein GAN loss for kinematic chain $j$ as:
\begin{equation} \label{eqn:chainWGANloss}
\begin{aligned}
\mathcal{L}_{\text{Chain}^j} = \mathop{\mathbb{E}}_{\mathbf{C}^j}\mathbf{D}^j[C^j_{t+1:t+T}] -  \mathop{\mathbb{E}}_{\mathbf{C'}^j}\mathbf{D}^j[C'^j_{t+1:t+T}]
\end{aligned}
\end{equation}
where $\mathbf{C}^j$ and $\mathbf{C}'^j$ respectively denote the real and generated future sequences for chain $j$.

On top of the Wasserstein GAN loss, we further introduce a stability loss given by the variance over $\mathcal{L}_{\text{Chain}^j}$.
\begin{equation} \label{eqn:stabilityloss}
\begin{aligned}
\mathcal{L}_{\text{Stability}} = \frac{1}{5}\sum_{j=1}^{5}(\mathcal{L}_{\text{Chain}^j}-\overline{\mathcal{L}_{\text{Chain}^j}})^2.
\end{aligned}
\end{equation}

The local or chain level loss for chain $j$ is given by a weighted sum of the Wasserstein GAN loss for the chain as well as the stability loss,
\begin{equation} \label{eqn:localloss}
\begin{aligned}
\mathcal{L}_{\text{Local}}^j = \alpha\mathcal{L}_{\text{Chain}^j}+\beta\mathcal{L}_{\text{Stability}}+\gamma\mathcal{L}_{\text{GroundTruth}}^j
\end{aligned}
\end{equation}
where $\alpha$, $\beta$, and $\gamma$ are weight hyperparameters. $\mathcal{L}_{\text{GroundTruth}}^j$ is the difference between ground truth of chain $j$ and its prediction.

{\bf Pose-level loss.} The aggregation layer aggregates and balances the five kinematic chains to give a generated pose sequence $\mathbf{P}'_{t+1:t+T}$. A global critic network $\mathbf{D}^G$ learns the authenticity of this pose, incurring a pose-level Wasserstein GAN loss as:
\begin{equation} \label{eqn:poseWGANloss}
\begin{aligned}
\mathcal{L}_{\text{Pose}} =
\mathop{\mathbb{E}}_{\mathbf{P}}\mathbf{D}^G[P_{t+1:t+T}] -  \mathop{\mathbb{E}}_{\mathbf{P'}}\mathbf{D}^G[P'_{t+1:t+T}]
\end{aligned}
\end{equation}
where $\mathbf{P}$ denotes the real future pose sequence.

To ensure temporal smoothness, a consistency loss is incorporated as
\begin{equation} \label{eqn:consistencyloss}
\begin{aligned}
\mathcal{L}_{\text{Consistency}} = \sum_{i=1}^T{|\mathbf{P}'_{t+i}-\mathbf{P}'_{t+i-1}|^2}.
\end{aligned}
\end{equation}

The global or pose-level loss for the global GAN is given by a weighted sum of the pose Wasserstein GAN loss and the consistency loss:
\begin{equation} \label{eqn:globalloss}
\begin{aligned}
\mathcal{L}_{\text{Global}} = \lambda\mathcal{L}_{\text{Pose}} + \mu\mathcal{L}_{\text{Consistency}} + \eta \mathcal{L}_{\text{GroundTruth}}
\end{aligned}
\end{equation}
$\mathcal{L}_{\text{GroundTruth}}$ is the difference between the ground truth pose sequence $\mathbf{P}$ and its prediction $\mathbf{P'}$.

\textbf{Action transfer.} The training for  action transfer is challenging since there are no public MoCap datasets involving transitions between different actions. We can only train in an unsupervised way. As illustrated in Fig. \ref{fig:mergeaction}, we perform interpolation in the latent space to fill in the discontinuity between two sequences.  Specifically, let $\mathbf{P}=(\mathbf{P}_{1},\cdots,\mathbf{P}_{t})$ be a sequence of action $A_1$ and $\mathbf{P}'=(\mathbf{P}'_{m+1}, \cdots, \mathbf{P}'_{m+T})$ a sequence of action $A_2$. We are to transfer from $\mathbf{P}$ to $\mathbf{P}'$. We may first train the VAE generator using large-scale ground truth sequences of all actions, making it able to predict the immediate future pose  (\textbf{not} future sequence). Then, feed $\mathbf{P}$ and  $(\mathbf{P}'_{m-t+1}, \cdots, \mathbf{P}'_{m})$ respectively to the trained VAE network, obtaining their latent vectors  $\mathbf{z}_1$ and $\mathbf{z}_2$. We perform interpolation between  $\mathbf{z}_1$ and $\mathbf{z}_2$ to fill in the discontinuity between $\mathbf{P}_{t}$ to $\mathbf{P}'_{m+1}$. Given that $\mathbf{z}_2$ contains motion contexts towards generating prediction $\mathbf{P}'_{m+1}$, the decoded interpolations will gradually transfer from $\mathbf{P}_{t}$ to $\mathbf{P}'_{m+1}$.

\begin{table*}[t]
	{
	\begin{tabular}{l|llll|llll|llll|llll}
		\hline
		Time  & 80 & 160 & 400 & 1000 & 80 & 160 & 400 & 1000 & 80 & 160 & 400 & 1000 & 80 & 160 &400 & 1000 \\ 
		\hline
		& \multicolumn{4}{c|}{Directions}  & \multicolumn{4}{c|}{Greeting}    & \multicolumn{4}{c|}{Phoning} & \multicolumn{4}{c}{Posing} \\ 
		\hline
		HMR  & 0.38 & 0.58 &0.90 & 1.43 & 0.52 & {0.85} & 1.40 & 1.73 & 0.56 & 1.09 & 1.61 & 1.80 & 0.24 & 0.53 & 1.42 & 2.50 \\ 
		QNet  & \textbf{0.36} & \textbf{0.57} & 0.90 & 1.42 & 0.52 & \textbf{0.81} & 1.38 & 1.73 & 0.58 & 1.12 & 1.60 & 1.82 & 0.26 & 0.56 & \textbf{1.41} & 2.56 \\ 
		Ours & \textbf{0.36} & \textbf{0.57} & \textbf{0.89} & \textbf{1.41} & \textbf{0.51} & 0.86& \textbf{1.36} & \textbf{1.70} & \textbf{0.54}  & \textbf{1.05} & \textbf{1.58} & \textbf{1.79} & \textbf{0.22} & \textbf{0.51}  & \textbf{1.41} & \textbf{2.48}\\ 
		\hline
		& \multicolumn{4}{c|}{Purchases}   & \multicolumn{4}{c|}{Sitting}  & \multicolumn{4}{c|}{Sittingdown}     & \multicolumn{4}{c}{Takingphoto} \\ \hline
		HMR  & 0.58 & \textbf{0.83} & 1.25 & 1.99 & 0.38 & 0.61 &  1.15 & 1.66 & 0.39  & 0.75 & 1.28  & 1.91 & \textbf{0.21} & 0.44 & 1.01 & 1.42 \\ 
		QNet & 0.59 & \textbf{0.83} & 1.24 & 2.37 & 0.37 & \textbf{0.58} & \textbf{1.11} & 1.62 & 0.37 & 0.73  & 1.23  & 1.94 & 0.24 & 0.48 & 1.02 & 1.43 \\ 
		Ours & \textbf{0.55} & 0.85 & \textbf{1.23} & \textbf{1.95} & \textbf{0.35} & 0.60 &  1.13 & \textbf{1.60} & \textbf{0.36}  & \textbf{0.72} & \textbf{1.20}  & \textbf{1.85} & 0.23 & \textbf{0.41} & \textbf{0.99} & \textbf{1.40} \\ 
		\hline
		& \multicolumn{4}{c|}{Waiting}     & \multicolumn{4}{c|}{Walkingdog} & \multicolumn{4}{c|}{Walkingtogether} & \multicolumn{4}{c}{Average} \\ \hline
		HMR & 0.27 & 0.59 & 1.33 & 2.21 & 0.55 & 0.87  & 1.36 & 1.84 & 0.26  & 0.50 & 0.71  & 1.21 & 0.39 & 0.69  & 1.46  & 1.79 \\ 
		QNet  & 0.29 & 0.58 & \textbf{1.03} & 2.38 & \textbf{0.53} & 0.89 & 1.34 & 1.97 & 0.24 &  \textbf{0.45} & \textbf{0.68} & 1.31 & 0.38 & 0.69& 1.47 & 1.82 \\ 
		Ours & \textbf{0.23} & \textbf{0.56} & 1.29 & \textbf{2.15} & \textbf{0.53} & \textbf{0.85} & \textbf{1.33} & \textbf{1.80} & \textbf{0.22} & \textbf{0.45} & 0.73 & \textbf{1.19} & \textbf{0.37} & \textbf{0.67} & \textbf{1.43} & \textbf{1.75} \\ 
		\hline
	\end{tabular}
	}
	\caption{Performance evaluation (in MAE) over 11 remaining action types on the H3.6m dataset.The unit of time is ms.}
   	\label{tab:otheractions}
\end{table*}

\begin{table}[t]
\centering
{
	\begin{tabular}{l|l|l|l|l|l}
    \hline
    Time (ms) & 80 & 160 & 320 & 400 & 1000 \\ 
    \hline
     & \multicolumn{5}{c}{Walking} \\ 
    \hline
    Remove Local Critic & 0.40 & 0.63 & 0.73 & 0.85 & 1.01 \\ 
    Remove Global Critic & 0.35 & 0.61 & 0.72 & 0.84 & 1.92 \\ 
    Remove sub-GANs & 0.90 & 1.10 & 1.61 & 1.79 & 2.01 \\ 
    AM-GAN (Complete) & 0.23 & 0.51 & 0.62 & 0.66 & 0.84 \\ \hline
     & \multicolumn{5}{c}{Discussion} \\ 
    \hline
    Remove Local Critic & 0.40 & 0.71 & 0.89 & 1.20 & 2.89 \\
    Remove Global Critic & 0.29 & 0.62 & 0.87 & 1.00 & 1.19 \\ 
    Remove sub-GANs & 2.20 & 2.51 & 2.68 & 2.71 & 3.20 \\ 
    AM-GAN (Complete) & 0.28 & 0.55 & 0.81 & 0.92 & 1.58 \\ \hline
    \end{tabular}
}
	\caption{Ablation studies on different components of our AM-GAN evaluated over `Walking' and `Discussion'.}
   	\label{tab:ablation}
\end{table}

\begin{figure*}[t]
	\centering
	\includegraphics[width=0.7\textwidth]{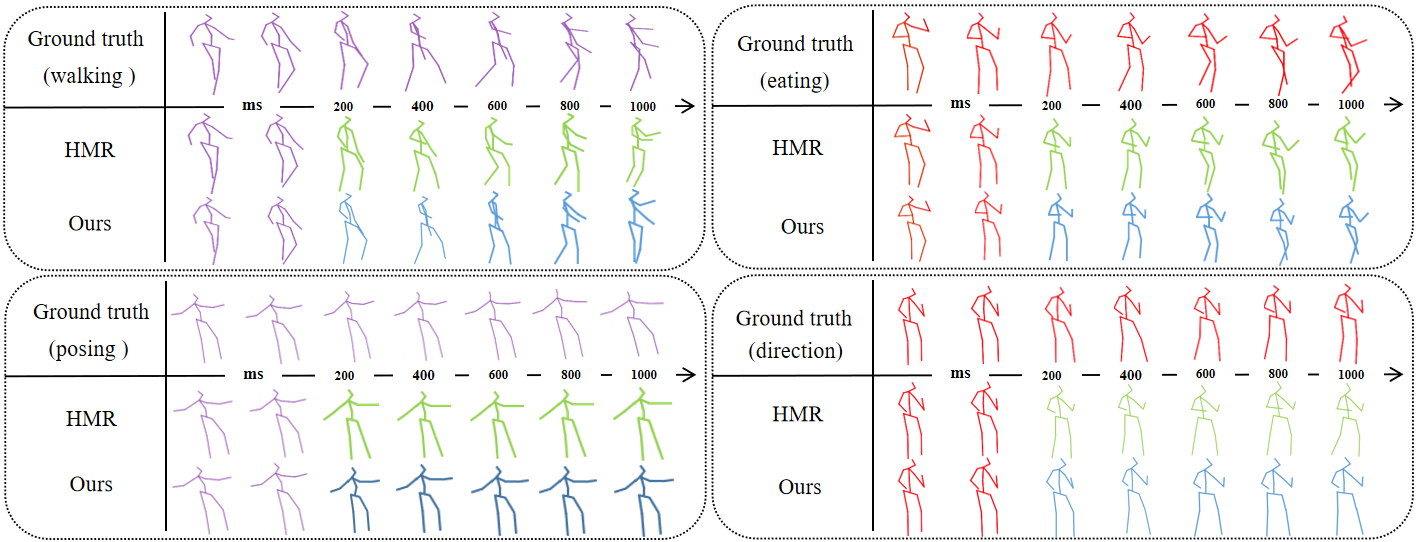}
	\caption{Some visual results.}
	\label{fig:Visual}
\end{figure*}

\begin{figure}[t]
	\centering
	\includegraphics[width=0.68\columnwidth]{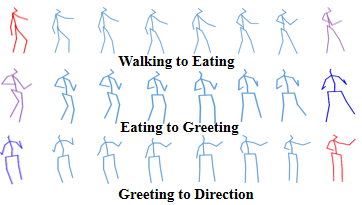}
	\caption{Action transfer examples.}	
	\label{fig:ActionTransfer}
\end{figure}

\begin{figure}[t]
	\centering
	\includegraphics[width=0.9\columnwidth]{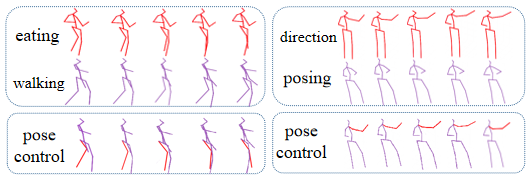}
	\caption{Fine-grained control over different chains in the predicted motion. In the left figure, we predict a `walking' action (purple) with the right leg following the action of `eating' (red). The result is shown on the bottom row with matching colors. In the right figure,  we predict a 'posing' action with the left hand following the `direction' action.}
		\label{fig:posecontrol}
\end{figure}

\textbf{Fine-grained pose control.} Breaking down the motion generation to chain level representations allows us to naturally incorporate fine-grained manipulation of each chain. Specifically, assuming that we are to generate the future sequence of an observed `walking' sequence $\textbf{P}$ and wants to control the right arm to act like `direction'. We can train the proposed network for each action independently. Then, we simply feed $\textbf{P}$ into the trained `walking' network to generate future `walking' action for the five chains. Similarly, we feed $\textbf{P}$ into the trained `direction' network to generate future `direction' action for the right arm chain. We replace the right arm chain generated by the `walking' network with that of the `direction' network, and feed the new five chains to the aggregation layer to generate a natural final pose sequence.

\section{Experiments}
\subsection{Dataset and Preprocessing}
In order to verify the ability of our model, we selected the largest and widely used data set: H3.6m \cite{H36M} that contains 3.6 million human images with 3D poses (comprising 25 distinct skeletal joints) obtained via the Vicon mocap system. There are 15 different action classes performed by 7 subjects. Following the protocol of previous methods \cite{erd,srnn}, subject 5 is used for testing while the data for the other 6 subjects are used for training. The observed sequence has 25 frames and the predicted output sequence is also set to 25 frames.

\subsection{Implementation Details}
We use Tensorflow to implement our method. For each of our sub-GAN, we have adopted a Variational  Auto-encoder (VAE) architecture \cite{kingma2014semi} 
as the generator module. The encoder and decoder of VAE consist of a GRU with 1,024 hidden units. The training of the GANs follows that of Wasserstein GANs \cite{gulrajani2017improved} but with our customized loss function outlined in the Loss Function section. The aggregation module is a full connected layer with 1,024 units. The local critic is a three-layer fully connected feedforward network with 512 units, while the global critic possesses 1,024 units. The Adam Optimizer \cite{kingma2014adam} is employed with initial learning 5e-5 and the batch size is set to 16. 

\subsection{Quantitative Evaluation on H3.6M Dataset}
We benchmark our method against previous methods, including ERD \cite{erd}, SRNN \cite{srnn}, RRNN \cite{residualgru}, LSTM-AE \cite{tu2018spatial}, HP-GAN \cite{barsoum2018hp}, CEM \cite{li2018convolutional}, BiHMP-GAN \cite{kundu2019bihmp}, SKel-Net \cite{guo2019human}, HMR \cite{liu2019towards} and QuaterNet \cite{pavllo2019modeling}. Following previous work, the mean angle error (MAE) is adopted as the evaluation metric. We report the results in Table \ref{tab:allmethods} over four actions `Walking', `Eating', `Smoking', `Discussion'.

As shown in Table \ref{tab:allmethods}, our method generally achieves superior performance in quantitative accuracy over the state-of-the-arts. One case in point is that the actions have different levels of complexity. For example, walking is a more regular action and demonstrates a high degree of periodicity whereas discussion is more complex and basically aperiodic. As would be expected, prediction accuracy falls with complexity of the action, which is observed throughout the empirical results. One noteworthy remark is that our method demonstrates significant improvements for the more complex action types such as discussion. This can be attributed to the fact that by modeling body parts and skeletal kinematic chains separately, our AM-GAN approach is less affected by lack of regularity and periodicity in the overall motion pattern. Furthermore, comparing against the other two GAN frameworks HP-GAN and BiHMP-GAN, the superior performance of our AM-GAN for both short-term and long-term prediction is also an indicator to the effectiveness of the divide and conquer approach in modeling local body parts and merging them at the final stage.

For the sake of completeness, we benchmark our method over the remaining 11 H3.6m actions against the two most recent and state-of-the-art works HMR and QuaterNet(QNet). The comparison results are shown in Table \ref{tab:otheractions}. Obviously,our method achieves new state-of-the-art performance quite consistently for short term and long term predictions. As shown in Fig.~\ref{fig:Visual}, we also visualize the results of our method against current excellent HMR, observing that our method delivers more natural and smooth sequences.

\subsection{Action Transfer}
We now consider scenarios where we control the specific action performed in the future. Some generated sequences depicting transitioning back and forth between different actions are showcased in Fig. \ref{fig:ActionTransfer}, where the middle six frames are generated transition frames. Empirical results show that our method is able to smoothly and authentically transfer from one action to another. 

\subsection{Human Pose Control}
We can assimilate high level controls over specific body parts with AM-GAN. The local and global level generation can be seamlessly combined to create natural and consistent motion sequences. As an example, we illustrate in the left figure of Fig. \ref{fig:posecontrol} on how we may control the movements of the right leg in the predicted sequence. Specifically, given a walking sequence, we generate its future `walking' sequence but enforce the right leg act like `eating' action prediction. Similarly, in the right figure of Fig.\ref{fig:posecontrol}, we show another example where we predict the future 'posing' action sequence but command the left hand to follow a `direction' action. 

\subsection{Ablation Studies}
We quantitatively analyze the contribution of various design components, including removing the 5 sub-GANs, removing the local critic and global critic. The empirical results are reported in Table \ref{tab:ablation}. One crucial finding is that incorporating sub-GANs and a divide and conquer strategy indeed improves over the baseline with a single GAN for modeling the motion generation. Modeling the constituent chains instead of the entire pose directly reduces complexity, thus improving the short-term and long-term prediction capabilities. Furthermore, it may be observed that the improvement over the baseline is more significant in the case of aperiodic actions such as `Discussion'. This is likely due to the fact that zooming in on individual body parts is more effective for modeling complex and irregular motion, where different body parts usually engage differently in the motion. Another observation is that the local critic enhances the short-term prediction whereas the global critic elevates long-term prediction performance.

\section{Conclusion}
In this paper, we propose a novel human motion prediction problem to incorporate control. In order to reduce the complexity of optimizing over the entire human pose manifold and to achieve fine-grained control, we employed a divide and conquer strategy and modeled each kinematic chain separately with our AM-GAN. On top of improving the state-of-the-art for both short and long-term predictions, AM-GAN also allows for controlling of the prediction spectrum. This includes generating transitions to a desired action or fine-grained control over specific body parts. We are positive that controllable motion generation would open many doors in adjacent fields.
\section{Acknowledgements}
This work was partly supported by the National Key Research and Development Program of China under No. 2020AAA0140004.   The Natural Science Foundation of Zhejiang Province, China No. LQ19F020001. The National Natural Science Foundation of China No. 61902348 and  No. 61772466. The Zhejiang Provincial Natural Science Foundation for Distinguished Young Scholars No. LR19F020003, and the Fundamental Research Funds for the Central Universities (Zhejiang University NGICS Platform).

\bigskip
\bibliography{References}

\end{document}